\documentclass[runningheads]{llncs}
\usepackage{graphicx}
\usepackage{balance}
\usepackage{amsmath}
\usepackage[tight,footnotesize]{subfigure}
\usepackage{bm}
\usepackage{algorithm}
\usepackage[noend]{algorithmic}
\usepackage{multirow}
\usepackage{amsfonts}

\begin{document}
\title{Beyond Submodularity: A Unified Framework of
Randomized Set Selection with Group Fairness Constraints}
\titlerunning{Randomized Set Selection with Group Fairness Constraints}
% If the paper title is too long for the running head, you can set
% an abbreviated paper title here
%
\author{ Shaojie Tang\inst{1}  \orcidID{0000-0001-9261-5210}  \and Jing Yuan\inst{2} \orcidID{0000-0001-6407-834X}}
\authorrunning{S. Tang and J. Yuan}
% First names are abbreviated in the running head.
% If there are more than two authors, 'et al.' is used.
%
\institute{Naveen Jindal School of Management, University of Texas at Dallas\\\email{shaojie.tang@utdallas.edu} \and Department of Computer Science and Engineering, University of North Texas \\\email{jing.yuan@unt.edu}  }
\maketitle              % typeset the header of the contribution
\begin{abstract}
Machine learning algorithms play an important role in a variety of important decision-making processes, including targeted advertisement displays, home loan approvals, and criminal behavior predictions. Given the far-reaching impact of these algorithms, it is crucial that they operate fairly, free from bias or prejudice towards certain groups in the population. Ensuring impartiality in these algorithms is essential for promoting equality and avoiding discrimination. To this end we introduce a unified framework for randomized subset selection that incorporates group fairness constraints.  Our problem involves a global utility function and a set of group utility functions for each group, here a group refers to a group of individuals (e.g., people) sharing the same attributes (e.g., gender). Our aim is to generate a distribution across feasible subsets, specifying the selection probability of each feasible set, to maximize the global utility function while meeting a predetermined quota for each group utility function in expectation. Note that there may not necessarily be any direct connections between the global utility function and each group utility function. We demonstrate that this framework unifies and generalizes many significant applications in machine learning and operations research. Our algorithmic results  either improves the best known result or provide the first approximation algorithms for new applications.
\end{abstract}
\section{Introduction}
The increasing use of machine learning algorithms in decision-making has raised concerns about the possibility of biases and discrimination. However, various efforts are being made to develop fair algorithms that ensure equitable outcomes for individuals or groups, even in sensitive domains.  These efforts involve creating techniques for classification \cite{zafar2017fairness}, ranking \cite{celis2017ranking}, clustering \cite{chierichetti2017fair}, bandit learning \cite{joseph2016fairness}, voting \cite{celis2018multiwinner}, college admission \cite{abdulkadirouglu2005college}, matching \cite{chierichetti2019matroids}, influence maximization \cite{tsang2019group}, and diverse data summarization \cite{el2020fairness}.

Various concepts of fairness have been proposed in the literature, including individual fairness, group fairness, and subgroup fairness. These concepts intend to tackle discrimination and bias in algorithmic decision-making, particularly in sensitive domains like employment, housing, and criminal justice. However, there is no universal measure of fairness since it often depends on the context and can be affected by various factors such as the decision-making process's objectives and the sensitive attribute in question. In this paper, we propose a general group fairness notation that unifies many notations from previous works. We assume there are $m$ groups defined by a set of shared attributes like race, gender, or age. To assess the appropriateness of a particular solution set $S$ for a specific group $t\in[m]$, we introduce $m$ group utility functions $g_1, g_2, \cdots, g_m: 2^V\rightarrow \mathbb{R}{\geq0}$. Each group utility function $g_t(S)$ evaluates the utility that group $t$ derives from the solution set $S$. For instance, in committee selection \cite{celis2018multiwinner}, $g_t(S)$ corresponds to the number of candidates chosen from group $t$ in the solution set $S$. Given a set of feasible subsets $\mathcal{F}$, let $x\in [0,1]^\mathcal{F}$ represent a distribution over solution sets in $\mathcal{F}$, where $x_S$ is the probability of choosing $S\in \mathcal{F}$. A distribution $x$ is deemed fair if $\sum{S\in \mathcal{F}}x_S g_t(S) \geq \alpha_t, \forall t \in[m]$, meaning the expected utility from each group $t\in[m]$ is lower bounded by $\alpha_t$. This enables the consideration of each group's representation in the final outcome, promoting diversity and preventing under-representation of any particular group.

 In addition, there is a global utility function $f:2^V\rightarrow \mathbb{R}_{\geq0}$. The expected (global) utility of a distribution $x$ can be computed as $\sum_{S\in \mathcal{F}}x_S f(S) $. Our goal is to find a distribution $x$ that maximizes $\sum_{S\in \mathcal{F}}x_S f(S) $ while ensuring that $\sum_{S\in \mathcal{F}}x_S g_t(S) \geq \alpha_t, \forall t \in[m]$. A formal definition of this problem is listed in $\textbf{P.0}$.  We show that this formulation brings together and extends numerous noteworthy applications in both machine learning and operations research. We next summarize the main contributions of this paper.
\begin{itemize}
\item We develop a polynomial-time algorithmic framework for $\textbf{P.0}$ based on ellipsoid method.
\item One main algorithmic finding (as formally presented in Theorem \ref{thm:2}) is that suppose that  for all  $z\in \mathbb{R}_{\geq 0}^m$, there exists a polynomial-time algorithm that  returns a set $A\in \mathcal{F}$ such that \[\forall S\in \mathcal{F}, f(A)+\sum_{t\in [m]}  g_t(A) \cdot z_t\geq \rho\cdot f(S)+\mu\cdot  \sum_{t\in [m]}  g_t(S) \cdot z_t,\] for some $\rho, \mu\in [0,1]$. Then there  is a polynomial-time $(\rho, \mu)$-approximation algorithm for  $\textbf{P.0}$. Here, the $\rho$ represents the approximation ratio, while the $\mu$ indicates that the solution may violate the fairness constraint by at most a factor of $\mu$. Hence, solving  $\textbf{P.0}$ is reduced to finding an approximation algorithm for a combinatorial subset selection problem. We utilize this result to tackle various applications such as fairness-aware submodular maximization, sequential submodular maximization with group constraints, and assortment planning with market share constraints. Our approach outperforms existing methods for some of these applications, while for others, we introduce novel applications and develop the first approximation algorithms.
\item Notably, when $f$ is non-negative monotone and submodular, and $g_t$ is a modular function, our approach gives a   \emph{feasible} and \emph{optimal} $(1-1/e, 1)$-approximation solution.
\item We explore extensions to other commonly used fairness metrics and propose effective algorithms for solving them. In the first extension, we introduce additional upper bounds $\beta_t$ on the expected utility of each group $t\in [m]$. Specifically, we require that $\alpha_t \leq \sum_{S\in \mathcal{F}}x_S\cdot g_t(S) \leq \beta_t , \forall t \in[m]$. In the second extension, we explore another frequently used measure of fairness that aims to achieve parity in pairwise utility between groups. The degree of fairness is defined by a parameter $\gamma$. Specifically, we require that for any two groups $t, t' \in[m]$, the difference between their expected utilities does not exceed $\gamma$, i.e., $\sum_{S\in \mathcal{F}}x_S\cdot g_t(S) - \sum_{S\in \mathcal{F}}x_S\cdot g_{t'}(S) \leq \gamma, \forall t, t' \in[m]$.
\end{itemize}

\paragraph{Additional Related Work.}  Over the years, there has been a significant effort to develop fair algorithms across various fields to address the issue of biased decision-making. In the domain of influence maximization and classification, researchers have been actively developing fair algorithms \cite{tsang2019group,zafar2017fairness}. Similarly, in voting systems, the focus has been on ensuring that election outcomes are a fair representation of the preferences of voters, leading to the development of fair algorithms \cite{celis2018multiwinner}. The field of bandit learning, which involves making sequential decisions based on uncertain information, has also seen a growing interest in the development of fair algorithms to address the issue of bias \cite{joseph2016fairness}. Additionally, the field of data summarization has seen an increasing focus on the development of fair algorithms \cite{celis2018fair} to provide a balanced representation of data and avoid biased decision-making.

The choice of fairness metric in previous studies depends on the context and type of bias being addressed, resulting in a range of optimization problems and fair algorithms customized to the particular demands of each application. We adopt a general group utility function to assess the solution's utility from each group's perspective.  Our framework is general enough to encompass numerous existing fairness notations, including the $80\%$-rule \cite{biddle2017adverse}, statistical parity \cite{10.1145/2090236.2090255}, and proportional representation \cite{monroe1995fully}. While most previous research on fairness-aware algorithm design (such as \cite{celis2018fair,yuan2023group,wang2021fair,mehrotra2021mitigating}) aims to find a deterministic solution set, our goal is to compute a randomized solution that can meet the group fairness constraints on average. This approach offers more flexibility in attaining group fairness. Our framework is general enough to encompass various existing studies on achieving group fairness through randomization, such as those examined in \cite{asadpour2022sequential,chen2022fair,tang2023ctw}.

\section{Preliminaries and Problem Statement}
We consider a set of $n$ items $V$ and $m$ groups. There is a \emph{global} utility function $f: 2^V\rightarrow \mathbb{R}_{\geq 0}$ and $m$ \emph{group} utility functions $g_1, g_2, \cdots, g_m:2^V\rightarrow \mathbb{R}_{\geq 0}$. Given a subset of items $S\subseteq V$, we use $g_t(S)$ to assess the utility of $S$ from each group $t$'s perspective. There is a required minimum expectation of utility from each group, represented by $\alpha\in \mathbb{R}_{\geq0}^{m}$, which acts as fairness constraints. This formulation enables us to design fair algorithms that take into account the preferences of each group, ensuring that the decision-making process is unbiased and leads to fair outcomes for all groups. Note that there may not be any connections between $f$ and $g_t$. A common way to define $g_t$, as elaborated on in Section \ref{sec:app1}, is by counting the number of items selected from group $t$.

 Suppose $\mathcal{F}$ contains all feasible subsets.  For example, if there exists a constraint that limits the selection of items to $k$, then $\mathcal{F}$ can be defined as $\{S\subseteq V\mid |S|\leq k\}$. The objective of our problem, denoted as $\textbf{P.0}$, is to find a distribution $x\in [0,1]^\mathcal{F}$ over $\mathcal{F}$ that maximizes the expected global utility, while also ensuring that the minimum expected utility from each group is met to comply with the fairness constraints. Here the decision variable $x_S\in[0, 1]$ specifies the selection probability of each feasible subset $S\in \mathcal{F}$. A formal definition of $\textbf{P.0}$ is listed in below.

 \begin{center}
\framebox[0.5\textwidth][c]{
\enspace
\begin{minipage}[t]{0.5\textwidth}
\small
$\textbf{P.0}$
$\max_{x\in [0,1]^\mathcal{F}} \sum_{S\in \mathcal{F}}x_S f(S)$ \\
\textbf{subject to:}
\begin{equation*}
\begin{cases}
\sum_{S\in \mathcal{F}}x_S\cdot g_t(S) \geq \alpha_t, \forall t \in[m].\\
\sum_{S\in \mathcal{F}}x_S\leq 1.
\end{cases}
\end{equation*}
\end{minipage}
}
\end{center}
\vspace{0.1in}

 This LP has $m+1$ constraints, excluding the standard constraints of $x_S\geq 0$ for all $S\in \mathcal{F}$. However, the number of variables in the LP problem is equivalent to the size of $\mathcal{F}$, which can become exponential in $n$. Due to this, conventional LP solvers are unable to solve this LP problem efficiently. %The next lemma asserts that $\textbf{P.0}$ is a problem that is NP-hard.

%\begin{lemma}
%Problem $\textbf{P.0}$ is NP-hard.
%\end{lemma}

\section{Approximation Algorithm for $\textbf{P.0}$}
\label{sec:2}

Before presenting our algorithm, we first introduce a combinatorial optimization problem called  \textsc{FairMax}. A solution to this problem serves as a subroutine of our final algorithm.
\begin{definition}[FairMax]
\label{def:min}
Given functions $f$ and $g_1, g_2, \cdots, g_m$,  a vector $z\in \mathbb{R}_{\geq 0}^m$ and a set of feasible subsets $\mathcal{F}$,  \textsc{FairMax}$(z, \mathcal{F})$ aims to
\begin{eqnarray}
\label{eq:yale}
\max_{S\in \mathcal{F}}(f(S)+\sum_{t\in [m]}  g_t(S) \cdot z_t).
\end{eqnarray}
\end{definition}

   I.e., \textsc{FairMax}$(z, \mathcal{F})$ seeks to find the feasible subset $S\in \mathcal{F}$ that maximizes $f(S)+\sum_{t\in [m]}  g_t(S) \cdot z_t$.

We next present the main theorem of this paper. A solution $y\in [0,1]^\mathcal{F}$ is said to achieve a $(a, b)$-approximation for $\textbf{P.0}$ if it satisfies the following conditions: $\sum_{S\in \mathcal{F}}y_S\leq 1$,  $\sum_{S\in \mathcal{F}}y_S f(S)\geq a\times OPT$, where $OPT$ denotes the optimal solution of $\textbf{P.0}$,   and $\sum_{S\in \mathcal{F}}y_S g_t(S) \geq b\times \alpha_t, \forall t \in[m]$.  Here, the $a$ represents the approximation ratio, while the $b$ indicates that the solution may violate the fairness constraint by at most a factor of $\mu$. The following theorem establishes a connection between \textsc{FairMax}$(z, \mathcal{F})$ and  $\textbf{P.0}$.

\begin{theorem}
\label{thm:2}
Suppose that  for all  $z\in \mathbb{R}_{\geq 0}^m$, there exists a polynomial-time algorithm that  returns a set $A\in \mathcal{F}$ such that \[\forall S\in \mathcal{F}, f(A)+\sum_{t\in [m]}  g_t(A) \cdot z_t\geq \rho\cdot f(S)+\mu\cdot  \sum_{t\in [m]}  g_t(S) \cdot z_t,\] for some $\rho, \mu\in [0,1]$. Then there  exists a polynomial-time $(\rho, \mu)$-approximation algorithm for  $\textbf{P.0}$.
\end{theorem}

To prove this theorem, it suffices to present a polynomial $(\rho, \mu)$-approximation algorithm for $\textbf{P.0}$, using a polynomial-time approximation algorithm for \textsc{FairMax}$(z, \mathcal{F})$ as a subroutine. To this end, we will investigate a relaxed form of $\textbf{P.0}$, which we refer to as $\textbf{RP.0}$.

 \begin{center}
\framebox[0.45\textwidth][c]{
\enspace
\begin{minipage}[t]{0.45\textwidth}
\small
$\textbf{RP.0}$
$\max_{x\in [0,1]^\mathcal{F}} \sum_{S\in \mathcal{F}}x_S f(S)$ \\
\textbf{subject to:}
\begin{equation*}
\begin{cases}
\sum_{S\in \mathcal{F}}x_S\cdot g_t(S) \geq \mu\alpha_t, \forall t \in[m].\\
\sum_{S\in \mathcal{F}}x_S\leq 1.
\end{cases}
\end{equation*}
\end{minipage}
}
\end{center}
\vspace{0.1in}

$\textbf{RP.0}$ is obtained by loosening the fairness constraint $\alpha_t$ in $\textbf{P.0}$ by a factor of $\mu\in[0,1]$, where $\mu$ is defined in Theorem \ref{thm:2}. By solving $\textbf{RP.0}$, we can obtain a solution that is approximately feasible for $\textbf{P.0}$. Given the assumptions made in Theorem \ref{thm:2}, in the following, we will focus on finding a solution for $\textbf{RP.0}$ and show that this solution constitutes a bicriteria $(\rho, \mu)$-approximation solution for the original problem $\textbf{P.0}$.

Note that the number of variables in $\textbf{RP.0}$ is equal to the number of elements in $\mathcal{F}$, which can become very large when $n$ is substantial. This results in standard LP solvers being unable to efficiently solve this LP problem. To address this challenge, we turn to the dual problem of $\textbf{RP.0}$ and use the ellipsoid algorithm \cite{grotschel1981ellipsoid} to solve it.

The dual problem of $\textbf{RP.0}$ (labeled as $\textbf{Dual of RP.0}$) involves assigning a weight $z_t\in \mathbb{R}_{\geq 0}$ to each group $t$ and introducing an additional variable, $w\in \mathbb{R}_{\geq 0}$.

  \begin{center}
   \framebox[0.7\textwidth][c]{
\enspace
\begin{minipage}[t]{0.7\textwidth}
\small
$\textbf{Dual of RP.0}$
$\min_{z\in \mathbb{R}_{\geq 0}^m,  w\in \mathbb{R}_{\geq 0}}  \sum_{t\in [m]} -\mu\alpha_t z_t+w$\\
\textbf{subject to:}
%\begin{equation*}
%\begin{cases}
%(1) \enspace D_i(t) = d_i \mathbf{I}_{[s_i, s_i + \tau_i]}(t)\\
%(1) \enspace D_t \triangleq  \sum_{i=1}^{n} d_i \mathbf{I}_{[s_i, s_i + \tau_i]}(t)\\
$ w \geq f(S)+\sum_{t\in [m]}  g_t(S) \cdot z_t, \forall S\in \mathcal{F}.$ %\\
 %z_t\geq 0, \forall t\in [m].\\
 % w \geq0.
%\sum_{i=1}^{|\mathcal{V}|} \min\{1,\sum_{j=1}^{|\mathcal{A}|} X_{ij}\} \leq P \mbox{ (\textbf{C3:} population constraint)}

%\end{cases}
%\end{equation*}
%\textbf{subject to:}
%\begin{equation*}
%\mathbf{x} \in \mathcal{X}
%\end{equation*}
\end{minipage}
}
\end{center}
\vspace{0.1in}

 Observe that the number of constraints in $\textbf{Dual of RP.0}$ might be exponential in $n$. At a high level, we aim to simplify this problem by reducing the number of constraints to a polynomial amount, without significantly altering the optimal solution.

We will now formally present our algorithm for $\textbf{Dual of RP.0}$  which involves a series of iterations known as the ellipsoid algorithm. During each iteration, the ellipsoid algorithm is used to determine whether the current solution is feasible or not by approximately solving an instance of \textsc{FairMax}. This problem acts as a test of feasibility and serves as a separation oracle, which helps to determine whether the current solution is located inside or outside the feasible region of the problem being solved.
 Let $C(L)$ denote the set of $(z\in \mathbb{R}_{\geq 0}^m, w\in \mathbb{R}_{\geq 0})$ satisfying that
\[\sum_{t\in [m]} -\mu\alpha_t z_t+w \leq L,\]
\[w \geq f(S)+\sum_{t\in [m]}  g_t(S) \cdot z_t, \forall S\in \mathcal{F}. \]

It is easy to verify that $L$ is achievable with respect to  $\textbf{Dual of RP.0}$ if and only if $C(L)$ is non-empty. To find the minimum value of $L$ such that $C(L)$ is non-empty, we use a binary search algorithm.

For a given $L$ and $(z, w)$, we first evaluate the inequality $\sum_{t\in [m]} -\mu\alpha_t z_t+w \leq L$.  If the inequality holds, the algorithm runs a subroutine $\mathcal{A}$ to solve \textsc{FairMax}$(z, \mathcal{F})$. Specifically, $\mathcal{A}$ aims to find the feasible subset $S\in \mathcal{F}$ that maximizes $f(S)+\sum_{t\in [m]}  g_t(S) \cdot z_t$. %Assuming that $\mathcal{A}$ is a $\mu$-approximation algorithm for \textsc{FairMax}$(z, \mathcal{F})$,
Let $A$ denote the set returned by $\mathcal{A}$.

\begin{itemize}
\item If the condition $f(A)+\sum_{t\in [m]} g_t(A) \cdot z_t \leq w$ holds, we mark $C(L)$ as a non-empty set (note that even in this case, $C(L)$ might still be empty because $A$ is only an approximate solution of \textsc{FairMax}$(z, \mathcal{F})$. However, as we will demonstrate later, this will not significantly impact our final solution). In such a scenario, we conclude that $L$ is achievable and proceed to try a smaller value of $L$.
\item If $f(A)+\sum_{t\in [m]} g_t(A) \cdot z_t > w$, this means that $(z, w)\notin C(L)$, and hence, $A$ is a separating hyperplane. To continue the optimization process, we search for a smaller ellipsoid with a center that satisfies this constraint. We repeat this process until we either find a feasible solution in $C(L)$, in which case we attempt a smaller $L$, or until the volume of the bounding ellipsoid becomes so small that it is considered empty with respect to $C(L)$. In the latter case, we conclude that the current objective is unattainable and will therefore try a larger $L$.
\end{itemize}

For a detailed understanding of the individual steps required to run ellipsoid with separation oracles and attain (multiplicative and additive) approximate guarantees, we recommend referring to Chapter 2 of \cite{bubeck2015convex}. After obtaining the results from the above ellipsoid methods, the subsequent procedures will encompass two primary steps. Firstly, an upper bound for the optimal solution of $\textbf{P.0}$ will be calculated. Following that, a $(\rho, \mu)$-approximation solution for $\textbf{P.0}$ will be computed.

\paragraph{Establishing an upper bound on the optimal solution of $\textbf{P.0}$.} Define $L^*$ to be the smallest value of $L$ for which $C(L)$ is marked as non-empty by our algorithm. We next show that the optimal solution of $\textbf{P.0}$  is at most  $ L^*/\rho$. To avoid trivial cases, let us assume that $\rho>0$.

Because $C(L^*)$ is marked as non-empty, there exists a $(z^*, w^*)$ such that
\begin{eqnarray}
\label{eq:11}
\sum_{t\in [m]} -\mu\alpha_t z^*_t+w^* \leq L^*
 \end{eqnarray}
 and
 \begin{eqnarray}
 \label{eq:12}
 f(A)+\sum_{t\in [m]} g_t(A) \cdot z^*_t \leq w^*.
 \end{eqnarray}

Given the assumption made regarding $A$ as stated in Theorem \ref{thm:2}, we have
\begin{eqnarray}
\label{eq:66}\forall S\in \mathcal{F}, f(A)+\sum_{t\in [m]}  g_t(A) \cdot z^*_t\geq \rho\cdot f(S)+\mu\cdot  \sum_{t\in [m]}  g_t(S) \cdot z^*_t.
 \end{eqnarray}

Because $\rho>0$, it follows that
\begin{eqnarray}
\label{eq:6}
\forall S\in \mathcal{F}, f(S)+\frac{\mu}{\rho}\cdot  \sum_{t\in [m]}  g_t(S) \cdot z^*_t \leq (f(A)+\sum_{t\in [m]}  g_t(A) \cdot z^*_t)/\rho   \leq w^*/\rho
 \end{eqnarray}
 where the first inequality follows from (\ref{eq:66}) and the second inequality is by  inequality (\ref{eq:12}). %Moreover, inequality (\ref{eq:11}) implies that
% \begin{eqnarray}
% \label{eq:7}
% \sum_{t\in [m]} -\alpha_t z^*_t+w^*/\mu \leq L^*/\mu.
% \end{eqnarray}

Consider the dual of $\textbf{P.0}$ (labeled as $\textbf{Dual of P.0}$), inequality (\ref{eq:6}) implies that $(\frac{\mu}{\rho}\cdot z^*, \frac{1}{\rho}\cdot w^*)$ is a feasible solution of $\textbf{Dual of P.0}$.
   \begin{center}
   \framebox[0.7\textwidth][c]{
\enspace
\begin{minipage}[t]{0.7\textwidth}
\small
$\textbf{Dual of P.0}$
$\min_{z\in \mathbb{R}_{\geq 0}^m,  w\in \mathbb{R}_{\geq 0}}
-\alpha_t z_t+w$\\
\textbf{subject to:}
$
%(1) \enspace D_i(t) = d_i \mathbf{I}_{[s_i, s_i + \tau_i]}(t)\\
%(1) \enspace D_t \triangleq  \sum_{i=1}^{n} d_i \mathbf{I}_{[s_i, s_i + \tau_i]}(t)\\
 w \geq f(S)+\sum_{t\in [m]}  g_t(S) \cdot z_t, \forall S\in \mathcal{F}. $
%\textbf{subject to:}
%\begin{equation*}
%\mathbf{x} \in \mathcal{X}
%\end{equation*}
\end{minipage}
}
\end{center}
\vspace{0.1in}

Plugging $(\frac{\mu}{\rho}\cdot z^*, \frac{1}{\rho}\cdot w^*)$   into the objective function of $\textbf{Dual of P.0}$, we can infer that the value of $\textbf{Dual of P.0}$  is at most $\sum_{t\in [m]} -\mu\alpha_t z^*_t/\rho+w^*/\rho \leq L^*/\rho$ where the inequality is by (\ref{eq:11}). By strong duality, the value of $\textbf{P.0}$  is at most  $ L^*/\rho$.  Hence, by finding a solution to $\textbf{RP.0}$  with a value of $L^*$, we can achieve an $(\rho, \mu)$-approximation for the original problem $\textbf{P.0}$.

\paragraph{Finding a solution to $\textbf{RP.0}$  with a value of $L^*-\epsilon$.}  Suppose $L^* - \epsilon$ is the largest value of $L$ for which the algorithm identifies that $C(L)$ is empty. Here, $\epsilon$ denotes the precision of the binary search.  We next focus on finding a solution to $\textbf{RP.0}$  with a value of $L^*-\epsilon$. To this end we can utilize only the feasible subsets from $\mathcal{F}$ that correspond to the separating hyperplanes obtained by the separation oracle. To achieve this, we define a subset of $\mathcal{F}$, denoted as $\mathcal{F}'$, which contains all the feasible subsets for which the dual constraint is violated during the implementation of the ellipsoid algorithm on $C(L^*-\epsilon)$. The size of $\mathcal{F}'$ is polynomial since the dual constraints are violated for only a polynomial number of feasible subsets. We can use the feasible subsets in $\mathcal{F}'$ to construct a polynomial sized dual linear program of $\textbf{RP.0}$ (labeled as $\textbf{Poly-sized Dual of P.0}$).

  \begin{center}
   \framebox[0.9\textwidth][c]{
\enspace
\begin{minipage}[t]{0.9\textwidth}
\small
$\textbf{Poly-zied Dual of RP.0}$
$\min_{z\in \mathbb{R}_{\geq 0}^m,  w\in \mathbb{R}_{\geq 0}}  \sum_{t\in [m]} -\mu\alpha_t z_t+w$\\
\textbf{subject to:}
%\begin{equation*}
%\begin{cases}
%(1) \enspace D_i(t) = d_i \mathbf{I}_{[s_i, s_i + \tau_i]}(t)\\
%(1) \enspace D_t \triangleq  \sum_{i=1}^{n} d_i \mathbf{I}_{[s_i, s_i + \tau_i]}(t)\\
$ w \geq f(S)+\sum_{t\in [m]}  g_t(S) \cdot z_t, \forall S\in \mathcal{F}'.$ %\\
 %z_t\geq 0, \forall t\in [m].\\
 % w \geq0.
%\sum_{i=1}^{|\mathcal{V}|} \min\{1,\sum_{j=1}^{|\mathcal{A}|} X_{ij}\} \leq P \mbox{ (\textbf{C3:} population constraint)}

%\end{cases}
%\end{equation*}
%\textbf{subject to:}
%\begin{equation*}
%\mathbf{x} \in \mathcal{X}
%\end{equation*}
\end{minipage}
}
\end{center}
\vspace{0.1in}

The objective of $\textbf{Poly-sized Dual of RP.0}$ is to maximize the dual objective function subject to the constraints defined by the feasible subsets in $\mathcal{F}'$. Because $C(L^*-\epsilon)$ is empty, the value of $\textbf{Poly-sized Dual of RP.0}$ at least $L^*-\epsilon$. Hence, the optimal solution to the dual of $\textbf{Poly-sized Dual of RP.0}$ (labeled as $\textbf{Poly-sized RP.0}$) is at least $L^*-\epsilon$.
 \begin{center}
\framebox[0.45\textwidth][c]{
\enspace
\begin{minipage}[t]{0.45\textwidth}
\small
$\textbf{Poly-sized RP.0}$
$\max_{x\in [0,1]^{\mathcal{F}'}} \sum_{S\in \mathcal{F}'}x_S f(S)$ \\
\textbf{subject to:}
\begin{equation*}
\begin{cases}
\sum_{S\in \mathcal{F}'}x_S\cdot g_t(S) \geq \mu\alpha_t, \forall t \in[m].\\
\sum_{S\in \mathcal{F}'}x_S\leq 1.
\end{cases}
\end{equation*}
\end{minipage}
}
\end{center}
\vspace{0.1in}

It is important to note that the size of $\textbf{Poly-sized RP.0}$ is polynomial, since $\mathcal{F}'$ contains only a polynomial number of feasible subsets. Thus, we can solve $\textbf{Poly-sized RP.0}$ efficiently and obtain a solution with a value of $L^*-\epsilon$.
This solution is a $(\rho, \mu)$-approximation (with additive error $\epsilon$) for $\textbf{P.0}$.

\section{Applications}
This section covers a range of applications for our framework, some of which yield better results than previously known methods. In other cases, we present new applications and provide the first approximation algorithms for them.

\subsection{Submodular Maximization with Group Fairness Constraints}
\label{sec:app1}
In this problem, we make the assumption that the global utility function $f$ and $m$ group utility functions $g_1, g_2, \cdots, g_m$ are submodular\footnote{A function $h: 2^V\rightarrow \mathbb{R}$ is considered submodular if, for any sets $X$ and $Y$ that are subsets of $V$ with $X\subseteq Y$ and any item $e\in V\setminus Y$, the following inequality holds: $h(X\cup\{e\})-h(X) \geq h(Y\cup\{e\})-h(Y)$. It is considered monotone if, for any set $X\subseteq V$ and any item $e \in V\setminus X$, it holds that
$h(X\cup\{e\})-h(X) \geq 0$.}. This problem setting is general enough to cover a wide range of optimization problems  that can be modeled using submodular global utility functions.  Examples of such applications include data summarization \cite{el2020fairness}, influence maximization \cite{kempe2008cascade}, and information retrieval \cite{yue2011linear}. It is worth noting that this scenario encompasses \emph{submodular maximization under submodular coverage} \cite{ohsaka2021approximation} as a special case. Their objective is to maximize a monotone submodular function subject to a lower quota constraint on a single submodular group utility function. However, their focus is on finding a deterministic solution set, whereas our approach provides greater flexibility in achieving group fairness. Next, we discuss the rationale behind assuming that the group utility function is submodular. In many prior works \cite{el2020fairness,celis2018multiwinner}, the concept of balance with respect to a sensitive attribute (such as race or gender) has been a widely used criterion for evaluating the solution obtained by fairness-aware optimization algorithms. This notion of balance typically involves ensuring that the solution does not significantly disadvantage any particular group with respect to the sensitive attribute, while also achieving good performance on the global objective $f$. We next provide a specific example in this context. Consider a set $V$ of $n$ items (such as people), where each item is associated with a sensitive attribute. Let $V_1, \dots, V_m$ denote the $m$ groups of items with the same attribute. We define a solution $x\in [0,1]^{\mathcal{F}}$, which encodes the selection probability of each set from $\mathcal{F}$,  to be fair if the expected number of selected items from each group $V_t$ is at least $\alpha_t$, where $\alpha_t$ often set proportional to the fraction of items of $V_t$ in the entire set $V$. In this case, we define $g_t(S)=|S \cap V_t|$ for each $t\in[m]$. It is easy to verify that $g_t$ is a monotone and submodular function.

Recall that solving  $\textbf{P.0}$ is reduced to solving \textsc{FairMax}$(z, \mathcal{F})$ (Definition \ref{def:min}). Here the objective of \textsc{FairMax}$(z, \mathcal{F})$ is to $\max_{S\in \mathcal{F}}(f(S)+\sum_{t\in [m]}  g_t(S) \cdot z_t)$. Fortunately, if $f$ and $g_t$ are submodular functions,  $f(S)+\sum_{t\in [m]}  g_t(S) \cdot z_t$ is also a submodular function by the fact that a linear combination of submodular functions is still submodular. If we assume that the family $\mathcal{F}$ is defined based on cardinality constraints, specifically as $\mathcal{F}=\{S \subseteq V\mid |S|\leq k\}$ for a positive integer $k$, then the problem of \textsc{FairMax}$(z, \mathcal{F})$ can be reduced to maximizing a submodular function while satisfying a cardinality constraint. According to \cite{nemhauser1978analysis}, if the objective function is non-negative, monotone and submodular, then an optimal $(1-1/e)$-approximation algorithm exists for this problem, that is, $\rho=\mu=1-1/e$. On the other hand, if the objective function is non-monotone (in addition to non-negative and submodular), then an approximation of $0.385$ is possible \cite{buchbinder2019constrained}, that is, $\rho=\mu=0.385$. This, together Theorem \ref{thm:2}, implies the following proposition.
\begin{proposition}
If $f$ and $g_t$ are non-negative monotone submodular functions, and $\mathcal{F}=\{S \subseteq V\mid |S|\leq k\}$ for a positive integer $k$, then there exists an optimal $(1-1/e, 1-1/e)$-approximation algorithm for $\textbf{P.0}$. If $f$ and $g_t$ are  non-negative  non-monotone submodular functions, there exists a  $(0.385, 0.385)$-approximation algorithm for $\textbf{P.0}$.
\end{proposition}
\subsubsection{Improved Results for Monotone Submodular $f$ and Modular $g_t$}
We next investigate an important special case of this application where we make the assumption that the global utility function $f$ is non-negative  monotone and submodular; $m$ group utility functions $g_1, g_2, \cdots, g_m$ are modular functions. It is easy to verify that the example previously discussed, wherein $g_t(S)=|S \cap V_t|$, satisfies the properties of a modular group utility function. We show that there exists a \emph{feasible} optimal  $(1-1/e)$-approximation algorithm for this special case.

Observe that if $f$ is non-negative  monotone and submodular, $g_t$ is a modular function (hence $\sum_{t\in [m]}  g_t(\cdot) \cdot z_t$ is also a modular function), and  $\mathcal{F}=\{S \subseteq V\mid |S|\leq k\}$ for a positive integer $k$, then \cite{sviridenko2017optimal} presented a randomized polynomial-time algorithm that produces a set $A\in \mathcal{F}$ such that for every $S\in \mathcal{F}$, it holds that
\begin{eqnarray}
\label{eq:hahaha}
f(A)+\sum_{t\in [m]}  g_t(A) \cdot z_t \geq (1-1/e)f(S)+\sum_{t\in [m]}  g_t(S) \cdot z_t.
\end{eqnarray}
That is, $\rho=1-1/e$ and $\mu=1$. Substituting these values into Theorem \ref{thm:2}, we have the following proposition. Note that  $\mu=1$ indicates that our solution strictly satisfies all fairness constraints.
\begin{proposition}
\label{cor:3}
If the global utility function $f$ is a non-negative monotone submodular function; $m$ group utility functions $g_1, g_2, \cdots, g_m$ are modular functions; $\mathcal{F}=\{S \subseteq V\mid |S|\leq k\}$ for a positive integer $k$, then there  exists  a \emph{feasible} $(1-1/e, 1)$-approximation algorithm for  $\textbf{P.0}$.
\end{proposition}

Note that the above result may be subject to a small additive error due to the omission of a similar error present in the original result (Inequality (\ref{eq:hahaha})) presented in \cite{sviridenko2017optimal}, which has been left out for simplicity.

\subsection{Sequential Submodular Maximization}
\label{sec:app2}
This problem was first studied in \cite{asadpour2022sequential} where the objective is to determine the optimal \emph{ordering} of a set of items to maximize a linear combination of various submodular functions. This variant of submodular maximization arises from the scenario where a platform displays a list of products to a user. The user examines the first $l$ items in the list, where $l$ is randomly selected from a given distribution and the user's decision to purchase an item from the set depends on a choice model, resulting in the platform's goal of maximizing the engagement of the shopper, which is defined as the probability of purchase. Formally, we are given monotone submodular functions $h_1, \cdots, h_n: 2^V\rightarrow \mathbb{R}_{\geq0}$ and $h^t_1, \cdots, h^t_n: 2^V\rightarrow \mathbb{R}_{\geq0}$ for each group $t\in[m]$, nonnegative coefficients $\lambda_1,\cdots, \lambda_n$ and $\lambda^t_1,\cdots, \lambda^t_n$
for each group $t\in[m]$. By abuse of notation, let $S$ be a permutation over items in $V$ and let $S_{[l]}$ represent the first $l$ items in $S$. Define the global utility function as $f(S) = \sum_{l \in [n]} \lambda_l h_l(S_{[l]})$.  In the context of product ranking, $\lambda_l$ represents the fraction of users with patience level $l$, while $h_l$ corresponds to the aggregate purchase probability function of users with patience level $l$. Similarly, the group utility function is defined as $g_t(S) = \sum_{l \in [n]} \lambda^t_l h^t_l(S_{[l]})$ for each group of users $t\in[m]$, where $\lambda^t_l$ should be interpreted as the fraction of users with patience level $l$ in group $t$. Despite $f$ and $g_t$ being defined over permutations instead of sets, it is easy to verify that Theorem \ref{thm:2} remains valid. In particular, suppose  for all  $z\in \mathbb{R}_{\geq 0}^m$, there exists a polynomial-time algorithm that  returns a permutation $A\in \mathcal{F}$ such that \[\forall S\in \mathcal{F}, f(A)+\sum_{t\in [m]}  g_t(A) \cdot z_t\geq \rho\cdot f(S)+\mu\cdot  \sum_{t\in [m]}  g_t(S) \cdot z_t,\] for some $\rho, \mu\in [0,1]$.  Here $\mathcal{F}$ is the set of all possible permutations over items in $V$. Then there  exists a polynomial-time $(\rho, \mu)$-approximation algorithm for the sequential submodular maximization problem.

Observe that in this case, the objective function of \textsc{FairMax}$(z, \mathcal{F})$  can be written as
\begin{eqnarray}
f(S)+\sum_{t\in [m]}  g_t(S) \cdot z_t &&= \sum_{l \in [n]} \lambda_l h_l(S_{[l]}) +\sum_{t\in [m]} (\sum_{l \in [n]} \lambda^t_l h^t_l(S_{[l]}) \cdot z_t )\\
&&=\sum_{l \in [n]} \lambda_l h_l(S_{[l]}) +\sum_{l \in [n]} ( \sum_{t\in [m]}  \lambda^t_l h^t_l(S_{[l]}) \cdot z_t )\\
&&=\sum_{l \in [n]} \left(\lambda_l h_l(S_{[l]}) + \sum_{t\in [m]}  \lambda^t_l h^t_l(S_{[l]}) \cdot z_t \right). \label{eq:4}
\end{eqnarray}
If $h_l$ and $h^t_l$ are both monotone and submodular for all $l\in[n]$ and $t\in[m]$, then $\lambda_l h_l(\cdot) + \sum_{t\in [m]}  z_t \lambda^t_l h^t_l(\cdot) $ is also monotone and submodular for all $l\in[n]$ by the fact that a linear combination of monotone submodular functions is still monotone and submodular.  According to the analysis presented in Theorem 1 of \cite{asadpour2022sequential}, if  $\lambda_l h_l(\cdot) + \sum_{t\in [m]}  z_t \lambda^t_l h^t_l(\cdot) $ is monotone and submodular for all $l\in[n]$, the problem of identifying a permutation $S$ that maximizes the right-hand side of equation (\ref{eq:4}) can be transformed into a submodular maximization problem subject to a (laminar) matroid constraint. Hence, there exists a $(1-1/e)$-approximation algorithm \cite{calinescu2007maximizing} for \textsc{FairMax}$(z, \mathcal{F})$, that is, $\rho=\mu=1-1/e$. Using  Theorem \ref{thm:2},  we obtain a $(1-1/e, 1-1/e)$-approximation algorithm for $\textbf{P.0}$. Note that the current state-of-the-art result  \cite{asadpour2022sequential} provides a bi-criteria $((1-1/e)^2, (1-1/e)^2)$-approximation for $\textbf{P.0}$. %Here, the first $(1-1/e)^2$ represents the approximation ratio, while the second $(1-1/e)^2$ indicates that the solution may violate the fairness constraint by at most a factor of $(1-1/e)^2$.
Our proposed framework offers significant improvements over their results in both approximation ratio and feasibility.

\begin{proposition}
There exists an optimal  $(1-1/e, 1-1/e)$-approximation algorithm for sequential submodular maximization.
\end{proposition}

\subsection{Random Assortment Planning with Group Market Share Constraints}
\label{sec:app3}
The third application concerns assortment planning, which is a problem that is widely recognized within the operations research community. Assortment planning with group market share constraints aims to identify the best possible combination of products to present to customers while ensuring that a minimum market share requirement of each group is met. Existing studies on this problem focus on finding a \emph{deterministic} solution that meets a minimum market share of a \emph{single} group. We extend this study to consider a randomized setting with multiple groups. Formally, this problem takes a set $V$ of $n$ products as input, which is divided into $m$ (possibly non-disjoint) groups denoted by $V_1, V_2, \cdots, V_m$. Under the well-known multinomial logit (MNL) model, each product $i\in V$ has a preference weight $\nu_i$ and let $\nu_0$ denote the preference for no purchase. Let $r_i$ denote the revenue of selling a product $i\in V$. Given an assortment $S\subseteq V$, the purchase probability of any product $i\in S$ is $\frac{ \nu_i}{\nu_0+\sum_{i\in S} \nu_i}$. Hence the expected revenue of offering $S$ is $f(S)=\frac{\sum_{i\in S} r_i \nu_i}{\nu_0+\sum_{i\in S}  \nu_i}$ and the resulting market share of group $t\in[m]$ is $g_t(S)=\frac{\sum_{i\in S\cap V_t} \nu_i}{\nu_0+\sum_{i\in S}  \nu_i}$. %Because there is only one group, we drop the subscript $t$ from both the function $g_t$ and the variable $\alpha_t$.
Assume $\mathcal{F}$  is comprised of all possible subsets of $V$, the goal of $\textbf{P.0}$ is to compute the selection probability $x_S$ of each assortment of products $S \subseteq V$  such that the expected revenue $\sum_{S\in \mathcal{F}}x_S f(S)$ is maximized while the expected market share is at least $\sum_{S\in \mathcal{F}}x_Sg_t(S)\geq \alpha_t$ for each group $t\in[m]$.  To solve this problem, we consider its corresponding \textsc{FairMax}$(z, \mathcal{F})$, whose objective is to find a $S\subseteq V$ that maximizes $f(S)+  \sum_{t\in[m]}g_t(S) \cdot z_t$ for a given vector $z\in \mathbb{R}^m_{\geq 0}$. By the definitions of $f$ and $g_t$, the objective function of \textsc{FairMax}$(z, \mathcal{F})$ can be written as
\begin{eqnarray}
f(S)+  \sum_{t\in[m]}g_t(S) \cdot z_t &&= \frac{\sum_{i\in S} r_i \nu_i}{\nu_0+\sum_{i\in S}  \nu_i} +\sum_{t\in[m]} \frac{\sum_{i\in S\cap V_t} \nu_i}{\nu_0+\sum_{i\in S}  \nu_i}\cdot z_t\\
&&=  \frac{\sum_{i\in S} (r_i+\sum_{t\in [m]}z_t\cdot \mathbf{1}_{i\in V_t}) \nu_i}{\nu_0+\sum_{i\in S}  \nu_i},
 \end{eqnarray}
where $\mathbf{1}_{i\in V_t}\in\{0, 1\}$ is an indicator variable such that $\mathbf{1}_{i\in V_t}=1$ if and only if $i\in V_t$.

Hence, the goal of \textsc{FairMax}$(z, \mathcal{F})$ is to
\begin{eqnarray}
\label{eq:3}
\max_{S\subseteq V}\frac{\sum_{i\in S} (r_i+\sum_{t\in [m]}z_t\cdot \mathbf{1}_{i\in V_t}) \nu_i}{\nu_0+\sum_{i\in S}  \nu_i}.
 \end{eqnarray}

This problem can be viewed as an unconstrained assortment planning problem, where each product $i\in V$ has a revenue of $(r_i+\sum_{t\in [m]}z_t\cdot \mathbf{1}_{i\in V_t})$ and a preference weight of $\nu_i$, and the preference weight of no purchase is $\nu_0$. According to \cite{talluri2004revenue}, the optimal solution for this problem is a revenue-ordered assortment. In other words, the assortment that maximizes revenue consists of the $l$ products with the highest revenues $(r_i+\sum_{t\in [m]}z_t\cdot \mathbf{1}_{i\in V_t})$, where $l\in[n]$. As a result, \textsc{FairMax}$(z, \mathcal{F})$ (problem (\ref{eq:3})) can be solved optimally in polynomial time by examining at most $n$ potential assortments. Thus, the combination of Theorem \ref{thm:2} and the ability to solve \textsc{FairMax}$(z, \mathcal{F})$ optimally in polynomial time (i.e., $\rho=\mu=1$) implies that an optimal solution for $\textbf{P.0}$ exists.

\begin{proposition}
There exists an optimal and feasible algorithm for assortment planning with group market share constraints.
\end{proposition}

\section{Discussion on Other Variants of Fairness Notations}
In this section, we examine two additional notations of fairness that are frequently employed in the literature.
\subsection{Incorporating  Fairness Upper Bound Constraints $\beta_t$}
One natural way to extend the fairness notation we introduced in $\textbf{P.0}$ would be to impose further upper bounds $\beta_t$ on the expected utility of every group. Formally,
\begin{center}
\framebox[0.6\textwidth][c]{
\enspace
\begin{minipage}[t]{0.6\textwidth}
\small
$\textbf{P.A}$
$\max_{x\in [0,1]^\mathcal{F}} \sum_{S\in \mathcal{F}}x_S f(S)$ \\
\textbf{subject to:}
\begin{equation*}
\begin{cases}
\alpha_t \leq \sum_{S\in \mathcal{F}}x_S\cdot g_t(S) \leq \beta_t , \forall t \in[m].\\
\sum_{S\in \mathcal{F}}x_S\leq 1.
\end{cases}
\end{equation*}
\end{minipage}
}
\end{center}
\vspace{0.1in}

This general formulation can be seen in several previous studies on fairness-aware optimization, such as  \cite{celis2018fair,el2020fairness}. Fortunately, we can still use ellipsoid method to solve $\textbf{P.A}$ to obtain a bicriteria  algorithm. As we will see later,   the following problem of  \textsc{FairMax} serves as a separation oracle, which helps to determine whether the current solution is located inside or outside the feasible region of the problem being solved.
\begin{definition}[FairMax]
\label{def:minA}
Given functions $f$ and $g_1, g_2, \cdots, g_m$,  two vectors $z\in \mathbb{R}_{\geq 0}^m$ and $u\in \mathbb{R}_{\geq 0}^m$, and a set of feasible subsets $\mathcal{F}$,  \textsc{FairMax}$(z, u, \mathcal{F})$ aims to
\begin{eqnarray}
\max_{S\in \mathcal{F}}( f(S)+\sum_{t\in [m]}  g_t(S) \cdot (z_t-u_t)).
\end{eqnarray}
\end{definition}

Unlike the objective function in (\ref{eq:yale}), the coefficient of $g_t$ in the above utility function might take on negative values. The next theorem builds a connection between  \textsc{FairMax}$(z, u, \mathcal{F})$ and $\textbf{P.A}$. Here we extend the definition of $(a, b)$-approximation such that a  solution $y\in \mathbb{R}_{\geq 0}^m$ is said to achieve a $(a, b)$-approximation for $\textbf{P.A}$ if it satisfies the following conditions: $\sum_{S\in \mathcal{F}}y_S\leq 1$,  $\sum_{S\in \mathcal{F}}y_S f(S)\geq a\times OPT$, where $OPT$ denotes the optimal solution of $\textbf{P.A}$,   and $\beta_t \geq \sum_{S\in \mathcal{F}}y_S g_t(S) \geq b\times \alpha_t, \forall t \in[m]$. The proof of the following theorem is moved to appendix.

%Following the same proof of Theorem \ref{thm:2}, except that $z_t$ is replaced by $(z_t-u_t)$, we have the following theorem.
   \begin{theorem}
\label{thm:21A}
Assuming  for all  $z\in \mathbb{R}_{\geq 0}^m$ and $u\in \mathbb{R}_{\geq 0}^m$,  there exists a polynomial-time algorithm that  returns a set $A\in \mathcal{F}$ such that \[\forall S\in \mathcal{F}, f(A)+\sum_{t\in [m]}  g_t(A) \cdot z_t\geq \rho\cdot f(S)+\mu\cdot  \sum_{t\in [m]}  g_t(S) \cdot (z_t-u_t),\] for some $\rho, \mu\in[0,1]$. Then there  exists a polynomial-time $(\rho, \mu)$-approximation algorithm for  $\textbf{P.A}$.
\end{theorem}

 Observe that if $f$ is non-negative  monotone and submodular, $g_t$ is a modular function (hence $\sum_{t\in [m]}  g_t(\cdot) \cdot (z_t-u_t)$ is also a modular function), and  $\mathcal{F}=\{S \subseteq V\mid |S|\leq k\}$ for a positive integer $k$, then \cite{sviridenko2017optimal} presented a randomized polynomial-time algorithm that produces a set $A\in \mathcal{F}$ such that for every $S\in \mathcal{F}$, it holds that
\begin{eqnarray}
\label{eq:hahahaA}
f(A)+\sum_{t\in [m]}  g_t(A) \cdot (z_t-u_t) \geq (1-1/e) f(S)+\sum_{t\in [m]}  g_t(S) \cdot (z_t-u_t).\end{eqnarray}

The following proposition follows immediately from Theorem \ref{thm:21A} and inequality (\ref{eq:hahahaA}). Note that this result recovers the findings from \cite{tang2023ctw}.
\begin{proposition}
\label{cor:4A}
If the global utility function $f$ is a non-negative monotone submodular function; $m$ group utility functions $g_1, g_2, \cdots, g_m$ are modular functions; $\mathcal{F}=\{S \subseteq V\mid |S|\leq k\}$ for a positive integer $k$, then there  exists  a \emph{feasible}  $(1-1/e, 1)$-approximation algorithm for  $\textbf{P.A}$.
\end{proposition}

For the case when $f$ is non-negative \emph{ non-monotone} and submodular, $g_t$ is a modular function, and $\mathcal{F}=\{S \subseteq V\mid |S|\leq k\}$ for a positive integer $k$, \cite{qi2022maximizing} presented an algorithm for \textsc{FairMax} that achieves $\rho=\frac{te^{-t}}{t+e^{-t}}-\epsilon$ and $\mu=\frac{t}{t+e^{-t}}$ for every constant $t\in[0,1]$. This, together with Theorem \ref{thm:21A}, indicates  a   $(\frac{te^{-t}}{t+e^{-t}}-\epsilon, \frac{t}{t+e^{-t}})$-approximation algorithm for  $\textbf{P.A}$.
\subsection{Pairwise Fairness}

We will now explore another frequently utilized notation of fairness, which relies on achieving parity in pairwise utility between groups. This type of notation has been employed in diverse scenarios, such as recommendation systems \cite{beutel2019fairness}, assortment planning \cite{chen2022fair}, ranking and regression models \cite{narasimhan2020pairwise}, and predictive risk scores \cite{kallus2019fairness}. Under this notion, it is expected that groups will experience comparable levels of utility. The extent of fairness is determined by a parameter $\gamma$. Specifically, we require that for every two groups  $t, t' \in[m]$, the difference between their expected utilities is at most $\gamma$, i.e.,  $\sum_{S\in \mathcal{F}}x_S\cdot g_t(S) - \sum_{S\in \mathcal{F}}x_S\cdot g_{t'}(S) \leq \gamma, \forall t, t' \in[m]$.  This problem is formally defined in $\textbf{P.B}$.
Our algorithmic findings extend to an even more general version of this problem by introducing a distinct $\gamma$ for each pair of groups. For the sake of simplicity, we do not elaborate on it here.
\begin{center}
\framebox[0.7\textwidth][c]{
\enspace
\begin{minipage}[t]{0.7\textwidth}
\small
$\textbf{P.B}$
$\max_{x\in [0,1]^\mathcal{F}} \sum_{S\in \mathcal{F}}x_S f(S)$ \\
\textbf{subject to:}
\begin{equation*}
\begin{cases}
\sum_{S\in \mathcal{F}}x_S\cdot g_t(S) - \sum_{S\in \mathcal{F}}x_S\cdot g_{t'}(S) \leq \gamma, \forall t, t' \in[m].\\
\sum_{S\in \mathcal{F}}x_S\leq 1.
\end{cases}
\end{equation*}
\end{minipage}
}
\end{center}
\vspace{0.1in}

In contrast to our approach for other fairness notations, we do not transform the original problem into a relaxed form for this particular case. Instead, we solve the dual of $\textbf{P.B}$ directly.  The dual of $\textbf{P.B}$ is presented in $\textbf{Dual of P.B}$.
%We first introduce a relaxed form of $\textbf{P.B}$ in $\textbf{RP.B}$ where the $\gamma$ constraint is replaced by $\gamma/\mu$ where $\mu\in[0, 1]$.
%
%\begin{center}
%\framebox[0.6\textwidth][c]{
%\enspace
%\begin{minipage}[t]{0.6\textwidth}
%\small
%$\textbf{RP.B}$
%$\max_{x\in [0,1]^\mathcal{F}} \sum_{S\in \mathcal{F}}x_S f(S)$ \\
%\textbf{subject to:}  \\
%\begin{equation*}
%\begin{cases}
%\sum_{S\in \mathcal{F}}x_S\cdot g_t(S) - \sum_{S\in \mathcal{F}}x_S\cdot g_{t'}(S) \leq \gamma/\mu, \forall t, t' \in[m].\\
%\sum_{S\in \mathcal{F}}x_S\leq 1.
%\end{cases}
%\end{equation*}
%\end{minipage}
%}
%\end{center}
%\vspace{0.1in}
%
%
%The dual of $\textbf{RP.B}$ is listed in $\textbf{Dual of RP.B}$.
%
%\begin{center}
%\framebox[0.6\textwidth][c]{
%\enspace
%\begin{minipage}[t]{0.6\textwidth}
%\small
%$\textbf{Dual of RP.B}$
%$\min_{z\in \mathbb{R}^{m \times m}, w\in \mathbb{R}_{\geq 0}} (\gamma/\mu) \sum_{t, t' \in[m]}z_{t, t'} + w $ \\
%\textbf{subject to:}
%$
%w \geq f(S)+\sum_{t, t' \in[m]}(g_{t'}(S)-g_t(S))z_{t, t'}, \forall S\in \mathcal{F}.$
%\end{minipage}
%}
%\end{center}
%\vspace{0.1in}

\begin{center}
\framebox[0.8\textwidth][c]{
\enspace
\begin{minipage}[t]{0.8\textwidth}
\small
$\textbf{Dual of P.B}$
$\min_{z\in \mathbb{R}_{\geq 0}^{m\times m}, w\in \mathbb{R}_{\geq 0}} \gamma \sum_{t, t' \in[m]}z_{t, t'} + w $ \\
\textbf{subject to:}
$
w \geq f(S)+\sum_{t, t' \in[m]}(g_{t'}(S)-g_t(S))z_{t, t'}, \forall S\in \mathcal{F}.$
\end{minipage}
}
\end{center}
\vspace{0.1in}

To solve $\textbf{Dual of P.B}$ using ellipsoid method, we define its separation oracle in \textsc{FairMax}.
\begin{definition}[FairMax]
\label{def:minB}
Given functions $f$ and $g_1, g_2, \cdots, g_m$,  a matrix $z\in \mathbb{R}_{\geq 0}^{m\times m}$ and a set of feasible subsets $\mathcal{F}$,  \textsc{FairMax}$(z, \mathcal{F})$ aims to
\begin{eqnarray}
\max_{S\in \mathcal{F}}(f(S)+\sum_{t, t' \in[m]}(g_{t'}(S)-g_t(S))z_{t, t'}).
\end{eqnarray}
\end{definition}

%Following the same proof of Theorem \ref{thm:2}, except that $z_t$ is replaced by $z_{t, t'}$ and $g_t(S)$ is replaced by $(g_{t'}(S)-g_t(S))$, we have the following theorem.
The next theorem builds a connection between  \textsc{FairMax}$(z, \mathcal{F})$ and $\textbf{P.B}$. In contrast to our results for other fairness notations, where we can only anticipate a bicriteria solution, we show that solving \textsc{FairMax}$(z, \mathcal{F})$  approximately leads to a \emph{feasible} solution for $\textbf{P.B}$. The proof of the following theorem is moved to appendix. %Here a solution $y\in \mathbb{R}_{\geq 0}^m$  achieves a $(a, b)$-approximation for $\textbf{P.B}$ if it satisfies the following conditions: $\sum_{S\in \mathcal{F}}y_S\leq 1$,  $\sum_{S\in \mathcal{F}}y_S f(S)\geq a\times OPT$, where $OPT$ denotes the optimal solution of $\textbf{P.B}$,   and $\sum_{S\in \mathcal{F}}x_S\cdot g_t(S) - \sum_{S\in \mathcal{F}}x_S\cdot g_{t'}(S) \leq \gamma/b, \forall t, t' \in[m]$.

   \begin{theorem}
\label{thm:211}
Suppose that  for all $z\in \mathbb{R}_{\geq 0}^{m\times m}$,  there exists a polynomial-time algorithm that  returns a set $A\in \mathcal{F}$ such that \[\forall S\in \mathcal{F}, f(A)+\sum_{t, t' \in[m]}(g_{t'}(A)-g_t(A))z_{t, t'}\geq \rho\cdot f(S)+\mu\cdot \sum_{t, t' \in[m]}(g_{t'}(S)-g_t(S))z_{t, t'},\] for some $\rho, \mu\in[0,1]$. Then there  exists a feasible  $\rho$-approximation algorithm for  $\textbf{P.B}$.
\end{theorem}

It should be noted that the above performance bound does not depend on $\mu$.  Observe that if $f$ is non-negative  monotone and submodular, $g_t$ is a modular function for all $t\in[m]$ (hence $\sum_{t, t' \in[m]}(g_{t'}(\cdot)-g_t(\cdot))z_{t, t'}$ is also a modular function for all $t, t'\in[m]$), and  $\mathcal{F}=\{S \subseteq V\mid |S|\leq k\}$ for a positive integer $k$, then \cite{sviridenko2017optimal} presented a randomized polynomial-time algorithm that produces a set $A\in \mathcal{F}$ such that, $\forall S\in \mathcal{F},$
\begin{eqnarray}
\label{eq:hahaha1}
 &&f(A)+\sum_{t, t' \in[m]}(g_{t'}(A)-g_t(A))z_{t, t'} ~\nonumber \\
 &&\geq (1-1/e) f(S)+\sum_{t, t' \in[m]}(g_{t'}(S)-g_t(S))z_{t, t'}.\end{eqnarray}

The following proposition follows immediately from Theorem \ref{thm:211} and inequality (\ref{eq:hahaha1}).
\begin{proposition}
\label{cor:4B}
If the global utility function $f$ is a non-negative monotone submodular function; $m$ group utility functions $g_1, g_2, \cdots, g_m$ are modular functions; $\mathcal{F}=\{S \subseteq V\mid |S|\leq k\}$ for a positive integer $k$, then there  exists  a feasible $(1-1/e)$-approximation algorithm for  $\textbf{P.B}$.
\end{proposition}

\textbf{Remark:} A recent work by \cite{chen2022fair} proposed an ellipsoid-based method for the assortment planning problem that includes pairwise fairness constraints. In page 14 of  \cite{chen2022fair} (the October 28th, 2022 version), they discussed the case where all items have uniform revenues. They showed that for this special case, their separation oracle is to maximize the summation of a non-negative monotone submodular function and a (not necessarily positive) modular function. They claimed that this objective function is a non-monotone submodular function and suggested using the continuous double greedy algorithm proposed in \cite{buchbinder2014submodular} to obtain a $[1/e+ 0.004, 1/2]$-approximation solution. However, it should be noted that the algorithm proposed by \cite{buchbinder2014submodular} only applies when the objective function is non-negative. Unfortunately, in general, the sum of a non-negative monotone submodular function and a (not necessarily positive) modular function can yield a negative value, rendering \cite{buchbinder2014submodular}'s algorithm inapplicable. On the other hand, our separation oracle in Definition \ref{def:minB} also has an objective function in this format, but using \cite{sviridenko2017optimal}'s algorithm as a subroutine to solve it leads to a $(1-1/e)$-approximation solution of our original problem. It is easy to verify that our framework (i.e., employing  \cite{sviridenko2017optimal}'s algorithm to solve the separation oracle) can be applied to the problem examined by \cite{chen2022fair} to obtain a $(1-1/e)$-approximation solution by conducting an analogous analysis to Theorem \ref{thm:211}.
\section{Conclusion}
In this paper, we introduce a general group fairness notation that unifies many notations used in previous works. We formulate the problem of finding a distribution over solution sets that maximizes global utility while satisfying fairness constraints. We develop a polynomial-time algorithmic framework based on the ellipsoid method to solve this problem. We also develop an optimal $(1-1/e)$-approximation algorithm for a special case of our problem, where $f$ is monotone and submodular, and $g_t$ is a modular function. This solution satisfies all fairness constraints strictly. Our work shows that this formulation brings together and extends numerous noteworthy applications in both machine learning and operations research.

\section{Appendix}
\subsection{Proof of Theorem \ref{thm:21A}} To prove this theorem, it suffices to present a polynomial $(\rho, \mu)$-approximation algorithm for $\textbf{P.A}$, using a polynomial-time approximation algorithm for \textsc{FairMax}$(z, u, \mathcal{F})$ as a subroutine. We first introduce a relaxed form of $\textbf{P.A}$ in $\textbf{RP.A}$ where the lower bound constraint is replaced with $\mu \alpha_t$.

\begin{center}
\framebox[0.6\textwidth][c]{
\enspace
\begin{minipage}[t]{0.6\textwidth}
\small
$\textbf{RP.A}$
$\max_{x\in [0,1]^\mathcal{F}} \sum_{S\in \mathcal{F}}x_S f(S)$ \\
\textbf{subject to:}
\begin{equation*}
\begin{cases}
\mu \alpha_t \leq \sum_{S\in \mathcal{F}}x_S\cdot g_t(S) \leq \beta_t , \forall t \in[m].\\
\sum_{S\in \mathcal{F}}x_S\leq 1.
\end{cases}
\end{equation*}
\end{minipage}
}
\end{center}
\vspace{0.1in}

The dual of $\textbf{RP.A}$ is listed in $\textbf{Dual of RP.A}$.

  \begin{center}
   \framebox[0.8\textwidth][c]{
\enspace
\begin{minipage}[t]{0.8\textwidth}
\small
$\textbf{Dual of RP.A}$
$\min_{z\in \mathbb{R}_{\geq 0}^m, u\in \mathbb{R}_{\geq 0}^m, w\in \mathbb{R}_{\geq 0}}  \sum_{t\in [m]}(\beta_tu_t -\mu \alpha_t z_t)+w$\\
\textbf{subject to:}
$
 w \geq f(S)+\sum_{t\in [m]}  g_t(S) \cdot (z_t-u_t), \forall S\in \mathcal{F}. $
%\textbf{subject to:}
%\begin{equation*}
%\mathbf{x} \in \mathcal{X}
%\end{equation*}
\end{minipage}
}
\end{center}
\vspace{0.1in}

Let $C(L)$ denote the set of $(z\in \mathbb{R}_{\geq 0}^m, u\in \mathbb{R}_{\geq 0}^m, w\in \mathbb{R}_{\geq 0})$ satisfying that
\[\sum_{t\in [m]}(\beta_tu_t -\mu \alpha_t z_t)+w\leq L,\]
\[w \geq f(S)+\sum_{t\in [m]}  g_t(S) \cdot (z_t-u_t), \forall S\in \mathcal{F}. \]

It is easy to verify that $L$ is achievable with respect to  $\textbf{Dual of RP.A}$ if and only if $C(L)$ is non-empty. To find the minimum value of $L$ such that $C(L)$ is non-empty, we use a binary search algorithm.

For a given $L$ and $(z, u, w)$, we first evaluate the inequality $\sum_{t\in [m]}(\beta_tu_t -\mu \alpha_t z_t)+w\leq L$.  If the inequality holds, the algorithm runs a subroutine $\mathcal{A}$ to solve \textsc{FairMax}$(z, u, \mathcal{F})$. Assuming that $\mathcal{A}$ is a $\mu$-approximation algorithm for \textsc{FairMax}$(z, u, \mathcal{F})$, let $A$ denote the set returned by $\mathcal{A}$.

\begin{itemize}
\item If the condition $ f(A)+\sum_{t\in [m]}  g_t(A) \cdot (z_t-u_t) \leq w$ holds, we mark $C(L)$ as a non-empty set. In such a scenario, we proceed to try a smaller value of $L$.
\item If $ f(A)+\sum_{t\in [m]}  g_t(A) \cdot (z_t-u_t) \cdot z_t > w$, this means that $(z, u, w)\notin C(L)$, and hence, $A$ is a separating hyperplane. We search for a smaller ellipsoid with a center that satisfies this constraint. We repeat this process until we either find a feasible solution in $C(L)$, in which case we attempt a smaller $L$, or until the volume of the bounding ellipsoid becomes so small that it is considered empty with respect to $C(L)$. In the latter case, we conclude that the current objective is unattainable and will therefore try a larger $L$.
\end{itemize}

 Define $L^*$ to be the smallest value of $L$ for which $C(L)$ is marked as non-empty by our algorithm.  We next show that the optimal solution of $\textbf{P.A}$  is at most  $ L^*/\rho$. To avoid trivial cases, let us assume that $\rho>0$.

Because $C(L^*)$ is marked as non-empty, there exists a $(z^*, u^*, w^*)$ such that
\begin{eqnarray}
\label{eq:11A}
\sum_{t\in [m]}(\beta_tu^*_t -\mu \alpha_t z^*_t)+w^*\leq L^*
 \end{eqnarray}
 and
 \begin{eqnarray}
 \label{eq:12A}
f(A)+\sum_{t\in [m]}  g_t(A) \cdot (z^*_t-u^*_t) \leq w^*.
 \end{eqnarray}

By the assumption made regarding $A$ in Theorem \ref{thm:21A}, we have $\forall S\in \mathcal{F},$
\begin{eqnarray}
\label{eq:6Ahaha} f(A)+\sum_{t\in [m]}  g_t(A) \cdot z^*_t\geq \rho\cdot f(S)+\mu\cdot  \sum_{t\in [m]}  g_t(S) \cdot (z^*_t-u^*_t). \end{eqnarray}

It follows that $\forall S\in \mathcal{F},$
\begin{eqnarray}
\label{eq:6A}
 &&f(S)+\frac{\mu}{\rho}\cdot \sum_{t\in [m]}  g_t(S) \cdot (z^*_t-u^*_t)  ~\nonumber \\
 && \leq (f(A)+\sum_{t\in [m]}  g_t(A) \cdot (z^*_t-u^*_t))/\rho   \leq w^*/\rho
 \end{eqnarray}
 where the first inequality follows from (\ref{eq:6Ahaha}) and the second inequality is by  inequality (\ref{eq:12A}). %Moreover, inequality (\ref{eq:11}) implies that
% \begin{eqnarray}
% \label{eq:7}
% \sum_{t\in [m]} -\alpha_t z^*_t+w^*/\mu \leq L^*/\mu.
% \end{eqnarray}

Consider the dual of $\textbf{P.A}$ (labeled as $\textbf{Dual of P.A}$), inequality (\ref{eq:6A}) implies that $(\frac{\mu}{\rho}\cdot z^*, \frac{\mu}{\rho}\cdot u^*,  \frac{1}{\rho}\cdot w^*)$ is a feasible solution of $\textbf{Dual of P.A}$.

  \begin{center}
   \framebox[0.8\textwidth][c]{
\enspace
\begin{minipage}[t]{0.8\textwidth}
\small
$\textbf{Dual of P.A}$
$\min_{z\in \mathbb{R}_{\geq 0}^m, u\in \mathbb{R}_{\geq 0}^m, w\in \mathbb{R}_{\geq 0}}  \sum_{t\in [m]}(\beta_tu_t -\alpha_t z_t)+w$\\
\textbf{subject to:}
$
 w \geq f(S)+\sum_{t\in [m]}  g_t(S) \cdot (z_t-u_t), \forall S\in \mathcal{F}. $
%\textbf{subject to:}
%\begin{equation*}
%\mathbf{x} \in \mathcal{X}
%\end{equation*}
\end{minipage}
}
\end{center}
\vspace{0.1in}

Plugging  $(\frac{\mu}{\rho}\cdot z^*, \frac{\mu}{\rho}\cdot u^*,  \frac{1}{\rho}\cdot w^*)$  into the objective function of $\textbf{Dual of P.A}$, we can infer that the value of $\textbf{Dual of P.A}$  is at most \begin{eqnarray}
\frac{\mu}{\rho}\cdot \sum_{t\in [m]}(\beta_tu^*_t -\alpha_t z^*_t)+w^*/\rho \leq \sum_{t\in [m]}(\beta_tu^*_t/\rho -\mu\alpha_t z^*_t/\rho)+w^*/\rho\leq L^*/\rho
 \end{eqnarray}
 where the first inequality is by the observations that $\beta_tu^*_t/\rho \geq 0$ and $\mu\in[0,1]$, and the second inequality is by (\ref{eq:11A}). By strong duality, the value of $\textbf{P.A}$  is at most  $ L^*/\rho$.  Hence, by finding a solution to $\textbf{RP.A}$  with a value of $L^*$, we can achieve an $(\rho, \mu)$-approximation for the original problem $\textbf{P.A}$.

 Suppose $L^* - \epsilon$ is the largest value of $L$ for which the algorithm identifies that $C(L)$ is empty. Here, $\epsilon$ denotes the precision of the binary search.  We next focus on finding a solution to $\textbf{RP.A}$  with a value of $L^*-\epsilon$. Define $\mathcal{F}'$ as the set that contains all the feasible subsets for which the dual constraint is violated during the implementation of the ellipsoid algorithm on $C(L^*-\epsilon)$. We use $\mathcal{F}'$ to construct a polynomial sized dual linear program of $\textbf{RP.A}$ (labeled as $\textbf{Poly-sized Dual of P.A}$).

  \begin{center}
   \framebox[1\textwidth][c]{
\enspace
\begin{minipage}[t]{1\textwidth}
\small
$\textbf{Poly-sizsed Dual of RP.A}$
$\min_{z\in \mathbb{R}_{\geq 0}^m, u\in \mathbb{R}_{\geq 0}^m, w\in \mathbb{R}_{\geq 0}}  \sum_{t\in [m]}(\beta_tu_t -\mu \alpha_t z_t)+w$\\
\textbf{subject to:}
$
 w \geq f(S)+\sum_{t\in [m]}  g_t(S) \cdot (z_t-u_t), \forall S\in \mathcal{F}'. $
%\textbf{subject to:}
%\begin{equation*}
%\mathbf{x} \in \mathcal{X}
%\end{equation*}
\end{minipage}
}
\end{center}
\vspace{0.1in}

 Because $C(L^*-\epsilon)$ is empty, the value of $\textbf{Poly-sized Dual of RP.A}$ at least $L^*-\epsilon$. Hence, the optimal solution to the dual of $\textbf{Poly-sized Dual of RP.A}$ (labeled as $\textbf{Poly-sized RP.A}$) is at least $L^*-\epsilon$.
\begin{center}
\framebox[0.6\textwidth][c]{
\enspace
\begin{minipage}[t]{0.6\textwidth}
\small
$\textbf{Poly-sized RP.A}$
$\max_{x\in [0,1]^{\mathcal{F}'}} \sum_{S\in \mathcal{F}'}x_S f(S)$ \\
\textbf{subject to:}
\begin{equation*}
\begin{cases}
\mu \alpha_t \leq \sum_{S\in \mathcal{F}'}x_S\cdot g_t(S) \leq \beta_t , \forall t \in[m].\\
\sum_{S\in \mathcal{F}'}x_S\leq 1.
\end{cases}
\end{equation*}
\end{minipage}
}
\end{center}
\vspace{0.1in}
Solving $\textbf{Poly-sized RP.A}$ obtains a solution with a value of $L^*-\epsilon$.
This solution is a $(\rho, \mu)$-approximation (with additive error $\epsilon$) for $\textbf{P.A}$.

\subsection{Proof of Theorem \ref{thm:211}} To prove this theorem, it suffices to present a feasible $\rho$-approximation algorithm for $\textbf{P.B}$, using a polynomial-time approximation algorithm for \textsc{FairMax}$(z, \mathcal{F})$ as a subroutine. Let $C(L)$ denote the set of $(z\in \mathbb{R}_{\geq 0}^{m\times m}, w\in \mathbb{R}_{\geq 0})$ satisfying that
\[\gamma \sum_{t, t' \in[m]}z_{t, t'} + w \leq L,\]
\[w \geq f(S)+\sum_{t, t' \in[m]}(g_{t'}(S)-g_t(S))z_{t, t'}, \forall S\in \mathcal{F}. \]

It is easy to verify that $L$ is achievable with respect to  $\textbf{Dual of P.B}$ if and only if $C(L)$ is non-empty. To find the minimum value of $L$ such that $C(L)$ is non-empty, we use a binary search algorithm.

For a given $L$ and $(z, w)$, we first evaluate the inequality $\gamma \sum_{t, t' \in[m]}z_{t, t'} + w \leq L$.  If the inequality holds, the algorithm runs a subroutine $\mathcal{A}$ to solve \textsc{FairMax}$(z, \mathcal{F})$. Let $A$ denote the set returned by $\mathcal{A}$.

\begin{itemize}
\item If the condition $ f(A)+\sum_{t, t' \in[m]}(g_{t'}(A)-g_t(A))z_{t, t'} \leq w$ holds, we mark $C(L)$ as a non-empty set. In such a scenario, we proceed to try a smaller value of $L$.
\item If $ f(A)+\sum_{t, t' \in[m]}(g_{t'}(A)-g_t(A))z_{t, t'} > w$, this means that $(z,w)\notin C(L)$, and hence, $A$ is a separating hyperplane. We search for a smaller ellipsoid with a center that satisfies this constraint. We repeat this process until we either find a feasible solution in $C(L)$, in which case we attempt a smaller $L$, or until the volume of the bounding ellipsoid becomes so small that it is considered empty with respect to $C(L)$. In the latter case, we conclude that the current objective is unattainable and will therefore try a larger $L$.
\end{itemize}

 Define $L^*$ to be the smallest value of $L$ for which $C(L)$ is marked as non-empty by our algorithm.  We next show that the optimal solution of $\textbf{P.B}$  is at most  $ L^*/\rho$. To avoid trivial cases, let us assume that $\rho>0$.

 Because $C(L^*)$ is marked as non-empty, there exists a $(z^*, w^*)$ such that
\begin{eqnarray}
\label{eq:11B}
\gamma\sum_{t, t' \in[m]}z^*_{t, t'} + w^* \leq L^*
 \end{eqnarray}
 and
 \begin{eqnarray}
 \label{eq:12B}
f(A)+\sum_{t, t' \in[m]}(g_{t'}(A)-g_t(A))z^*_{t, t'} \leq w^*.
 \end{eqnarray}

 By the assumption made regarding $A$ in Theorem \ref{thm:211}, we have $\forall S\in \mathcal{F},$
 \begin{eqnarray}
\label{eq:6Bhahah}
 &&f(A)+\sum_{t, t' \in[m]}(g_{t'}(A)-g_t(A))z^*_{t, t'} ~\nonumber \\
 &&\geq \rho\cdot f(S)+\mu\cdot \sum_{t, t' \in[m]}(g_{t'}(S)-g_t(S))z^*_{t, t'}.
 \end{eqnarray}

It follows that $\forall S\in \mathcal{F},$
\begin{eqnarray}
\label{eq:6B}
 &&f(S)+\frac{\mu}{\rho}\cdot \sum_{t, t' \in[m]}(g_{t'}(S)-g_t(S))z^*_{t, t'}   ~\nonumber \\
 &&\leq (f(A)+\sum_{t, t' \in[m]}(g_{t'}(A)-g_t(A))z^*_{t, t'} )/\rho   \leq w^*/\rho
 \end{eqnarray}
 where the first inequality follows from (\ref{eq:6Bhahah}) and the second inequality is by  inequality (\ref{eq:12B}). %Moreover, inequality (\ref{eq:11}) implies that
% \begin{eqnarray}
% \label{eq:7}
% \sum_{t\in [m]} -\alpha_t z^*_t+w^*/\mu \leq L^*/\mu.
% \end{eqnarray}

%Consider the dual of $\textbf{P.B}$ (labeled as $\textbf{Dual of P.B}$),
Inequality (\ref{eq:6B}) implies that $(\frac{\mu}{\rho}\cdot z^*, \frac{1}{\rho}\cdot w^*)$ is feasible for $\textbf{Dual of P.B}$.

\begin{center}
\framebox[0.8\textwidth][c]{
\enspace
\begin{minipage}[t]{0.8\textwidth}
\small
$\textbf{Dual of P.B}$
$\min_{z\in \mathbb{R}_{\geq 0}^{m\times m}, w\in \mathbb{R}_{\geq 0}} \gamma \sum_{t, t' \in[m]}z_{t, t'} + w $ \\
\textbf{subject to:}
$
w \geq f(S)+\sum_{t, t' \in[m]}(g_{t'}(S)-g_t(S))z_{t, t'}, \forall S\in \mathcal{F}.$
\end{minipage}
}
\end{center}
\vspace{0.1in}

%
%\begin{center}
%\framebox[0.6\textwidth][c]{
%\enspace
%\begin{minipage}[t]{0.6\textwidth}
%\small
%$\textbf{Dual of P.B}$
%$\min_{z\in \mathbb{R}^{m \times m}, w\in \mathbb{R}_{\geq 0}} \gamma \sum_{t, t' \in[m]}z_{t, t'} + w $ \\
%\textbf{subject to:}
%$
%w \geq f(S)+\sum_{t, t' \in[m]}(g_{t'}(S)-g_t(S))z_{t, t'}, \forall S\in \mathcal{F}.$
%\end{minipage}
%}
%\end{center}
%\vspace{0.1in}
Plugging  $(\frac{\mu}{\rho}\cdot z^*, \frac{1}{\rho}\cdot w^*)$ into the objective function of $\textbf{Dual of P.B}$, we can infer that the value of $\textbf{Dual of P.B}$  is at most \begin{eqnarray}
 \mu\gamma \sum_{t, t' \in[m]}(z^*_{t, t'}/\rho) + w^*/\rho \leq   \gamma \sum_{t, t' \in[m]}(z^*_{t, t'}/\rho) + w^*/\rho  \leq L^*/\rho
 \end{eqnarray}
 where the first inequality is by the observations that $\gamma\geq0, z^*_{t, t'}/\rho\geq 0, \mu\in[0,1]$, and the second inequality is by (\ref{eq:11B}). By strong duality, the value of $\textbf{P.B}$  is at most  $ L^*/\rho$.  Hence, by finding a solution to $\textbf{P.B}$  with a value of $L^*$, we can achieve a $\rho$-approximation for $\textbf{P.B}$.

 Suppose $L^* - \epsilon$ is the largest value of $L$ for which the algorithm identifies that $C(L)$ is empty. Define $\mathcal{F}'$ as the set that contains all the feasible subsets for which the dual constraint is violated during the implementation of the ellipsoid algorithm on $C(L^*-\epsilon)$. We use $\mathcal{F}'$ to construct a polynomial sized dual linear program of $\textbf{P.B}$ (labeled as $\textbf{Poly-sized Dual of P.B}$).

\begin{center}
\framebox[0.8\textwidth][c]{
\enspace
\begin{minipage}[t]{0.8\textwidth}
\small
$\textbf{Poly-sized Dual of P.B}$
$\min_{z\in \mathbb{R}^{m \times m}, w\in \mathbb{R}_{\geq 0}} \gamma \sum_{t, t' \in[m]}z_{t, t'} + w $ \\
\textbf{subject to:}
$
w \geq f(S)+\sum_{t, t' \in[m]}(g_{t'}(S)-g_t(S))z_{t, t'}, \forall S\in \mathcal{F}'.$
\end{minipage}
}
\end{center}
\vspace{0.1in}

 Because $C(L^*-\epsilon)$ is empty, the value of $\textbf{Poly-sized Dual of P.B}$ at least $L^*-\epsilon$. Hence, the optimal solution to the dual of $\textbf{Poly-sized Dual of P.B}$ (labeled as $\textbf{Poly-sized P.B}$) is at least $L^*-\epsilon$.
\begin{center}
\framebox[0.7\textwidth][c]{
\enspace
\begin{minipage}[t]{0.7\textwidth}
\small
$\textbf{Poly-sized P.B}$
$\max_{x\in [0,1]^{\mathcal{F}'}} \sum_{S\in \mathcal{F}}x_S f(S)$ \\
\textbf{subject to:}
\begin{equation*}
\begin{cases}
\sum_{S\in \mathcal{F}'}x_S\cdot g_t(S) - \sum_{S\in \mathcal{F}'}x_S\cdot g_{t'}(S) \leq \gamma, \forall t, t' \in[m].\\
\sum_{S\in \mathcal{F}'}x_S\leq 1.
\end{cases}
\end{equation*}
\end{minipage}
}
\end{center}
\vspace{0.1in}
Solving $\textbf{Poly-sized P.B}$ obtains a solution with a value of $L^*-\epsilon$.
This solution is a feasible $\rho$-approximation (with additive error $\epsilon$) for $\textbf{P.B}$.

\bibliographystyle{splncs04}
\bibliography{reference}
\end{document}